\documentclass[10pt,twocolumn,letterpaper]{article}

\usepackage{iccv}
\usepackage{times}
\usepackage{epsfig}
\usepackage{graphicx}
\usepackage{amsmath}
\usepackage{amssymb}
\usepackage{booktabs}       
\usepackage{multirow}
\usepackage{multicol}
\usepackage{algorithm}
\usepackage{algorithmicx}
\usepackage{algpseudocode}
\usepackage{verbatim}
\usepackage{subfig}
\usepackage{caption}
\usepackage{xcolor}

\usepackage[pagebackref=true,breaklinks=true,letterpaper=true,colorlinks,bookmarks=false]{hyperref}

\iccvfinalcopy 

\ificcvfinal\fi
\begin{document}

\title{Interpretable Disentanglement of Neural Networks by \\ Extracting Class-Specific Subnetwork}

\author{Yulong Wang \hspace{1.2cm} Xiaolin Hu \hspace{1.2cm} Hang Su\footnotemark \\
Institute for Artificial Intelligence, the State Key Laboratory of Intelligent Technology and Systems \\
Beijing National Research Center for Information Science and Technology, BNRist Lab\\
Department of Computer Science and Technology, Tsinghua University \\
{\tt\small \{wang-yl15@mails,xlhu@mail,suhangss@mail\}.tsinghua.edu.cn}
}
\maketitle

\begin{abstract}
    We propose a novel perspective to understand deep neural networks in an interpretable disentanglement form. For each semantic class, we extract a class-specific functional subnetwork from the original full model, with compressed structure while maintaining comparable prediction performance. The structure representations of extracted subnetworks display a resemblance to their corresponding class semantic similarities. We also apply extracted subnetworks in visual explanation and adversarial example detection tasks by merely replacing the original full model with class-specific subnetworks. Experiments demonstrate that this intuitive operation can effectively improve explanation saliency accuracy for gradient-based explanation methods, and increase the detection rate for confidence score-based adversarial example detection methods.
\end{abstract}

\section{Introduction}
Deep neural networks have recently transformed many areas including visual perceptions, language understanding, reinforcement learning, etc. Though they become the most representative intelligent systems with a dominant performance, DNNs are criticized for lacking transparency and interpretability. Better understanding the working mechanism of machine learning systems has become a requested demand, which is not only beneficial to academic research but also significant to many critical industries requiring a high level of safety concerns.

\begin{figure}[!t]
    \centering
    \includegraphics[width=0.8\columnwidth]{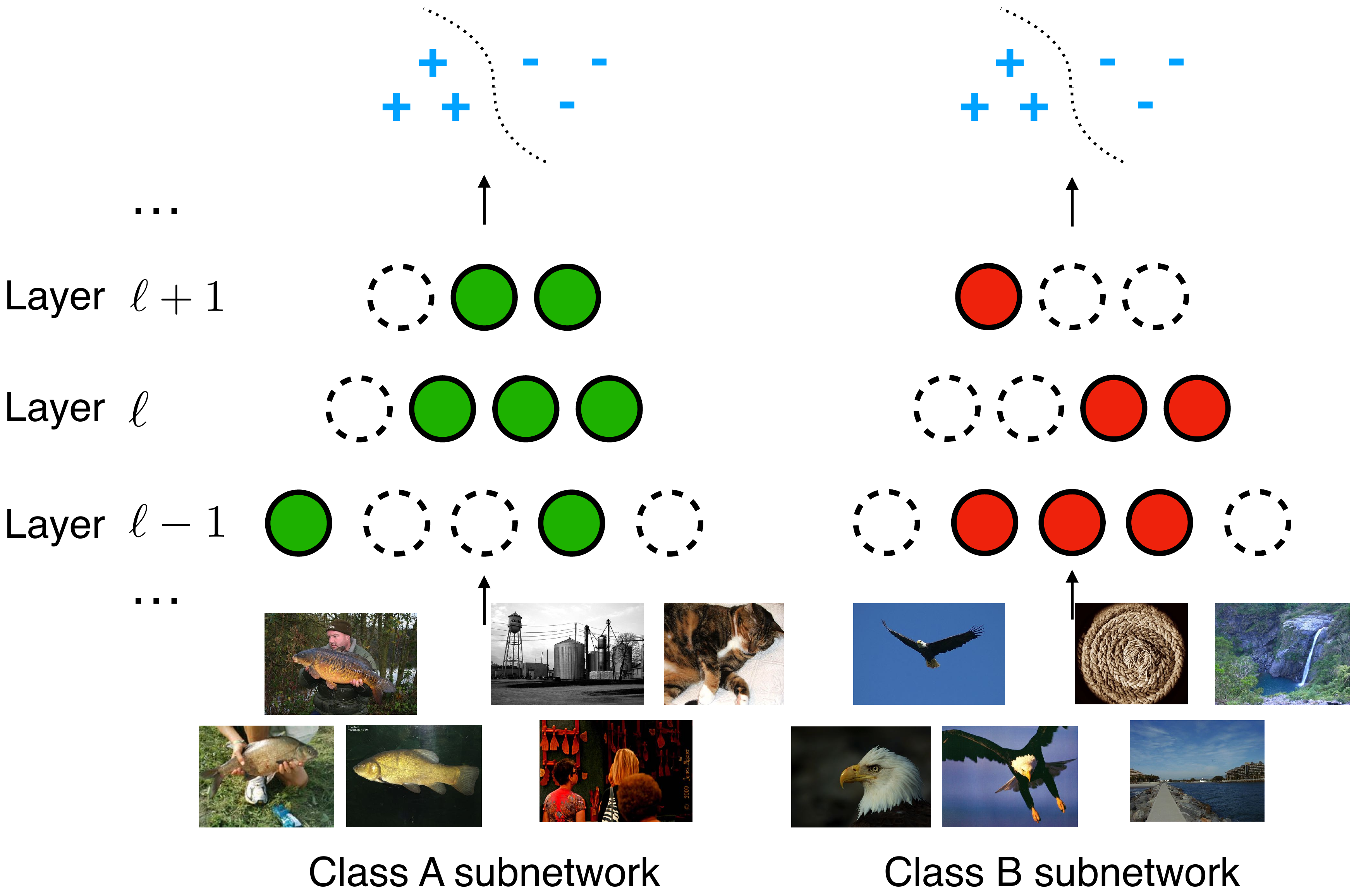}
    \caption{Method overview. For each class, we extract a subnetwork from the full model by learning to activate only a fraction of neurons on each layer. The extracted class-specific subnetwork can focus on one class prediction, and maintain comparable performance with the full model.}
    \vspace{-2ex}
    \label{fig:overview} 
\end{figure}

In this paper, we propose a simple and interpretable disentanglement form for deep neural networks, which can not only reveal neural network's functional behaviors but also have application improvement in visual explanation task~\cite{simonyan2013deconv} and adversarial example detection~\cite{feinman2017detecting}. The main idea is that we propose to extract the class-specific subnetwork for each semantic category from a pre-trained full model while maintaining a comparable prediction performance (Figure~\ref{fig:overview}). To effectively extract the subnetworks, we utilize the knowledge distillation criteria~\cite{hinton2015distilling} and model pruning strategy~\cite{liu2017learning}. We observe that the highly compressed subnetworks can display an architecture resemblance to their corresponding categorical semantic similarities.    

Furthermore, the interpretable subnetworks extracted by the proposed method can have further operational influence in other tasks. One application is visual explanation, which provides salient regions or features relevant to the model prediction results. We propose a simple improvement operation for gradient-based visual explanation method, by just replacing the full model weights with subnetwork for the requested explanation class. This can lead to more accurate and concise salient regions. The similar technique can also be applied to another application: adversarial sample detection, which detects the malicious samples fooling DNN classifiers. We propose to use the features generated by the class-specific subnetworks to construct confidence score-based detector since the resulting features are observed to be more separable from adversarial data than those generated by full models.

\section{Method}
Let $f_\theta(\cdot)$ denote a neural network with pre-trained weights $\theta$. For one test sample $x$, the output vector of $f_\theta(x)$ is then passed to the softmax activation function, which generates the prediction probability for each class. The output $f_\theta(x)$ for class $c$, which is usually termed as logit, can be expressed as $f^c_\theta(x)$, and $\boldsymbol{x}$ denotes a collection of samples $\{x_i\}$.

To extract class-specific subnetworks, we consider our problem under the knowledge distillation framework~\cite{hinton2015distilling}. Specifically, let $p(y|x;\theta)$ denotes the original prediction made by the full model parametrized by $\theta$ 
for a single sample $x$, and we want to extract a subnetwork with the parameter $\theta_c$ for class $c$, whose prediction $q(y|x;\theta_c)$ should be close to the full model
under KL divergence measurement. Therefore, the objective is 
\begin{equation}\label{eq:kl}
    \mathcal{L}(\theta_c|x) =  \mathbb{KL}\big(p(y|x;\theta)||q(y|x;\theta_c)\big) + \Omega(\theta_c),
\end{equation}
where $\Omega(\cdot)$ is an extra regularization term which encourages $\theta_c$ to be sparse enough. We adopt $\ell_1$-norm as regularization term.

We observe that the original full model already has a good performance for single class binary classification. Therefore, the probability can be represented by transforming the output logit into $p(y=c|x;\theta) = \sigma(f^c_\theta(x))$, where $\sigma$ is Sigmoid function. Then the objective in Equation~\eqref{eq:kl} can be rewritten as
\begin{align}
    \mathcal{L}(\theta_c |x) &=-\mathbb{E}_{p(y|x;\theta)}q(y|x;\theta_c) + \Omega(\theta_c) + \mathrm{const}\cr
    &=\mathrm{BCE}(f^c_\theta(x), f^c_{\theta_c}(x)) +\Omega(\theta_c),
\end{align}
where $\mathrm{BCE}$ stands for binary cross entropy function which is $\mathrm{BCE}(a,b) = - a\log b - (1-a)\log(1-b)$. By using Monte Carlo approximation, we can obtain the final objective for learning subnetwork $\theta_c$, which is
\begin{equation}\label{eq:final}
    \mathcal{L}(\theta_c) = \int_{x\in\mathcal{D}} p(x)\mathcal{L}(\theta_c|x)\mathrm{d}x\approx\frac{1}{|\mathcal{D}|}\sum_{x\in\mathcal{D}}\mathcal{L}(\theta_c|x). 
\end{equation}

As for the parametrization form of $\theta_c$, we associate control gates on multiple layers' output channels in the network. The control gate values $\lambda_i$ then modulate the output features of $i$-th layer by channel-wise multiplication. 

\section{Adversarial Sample Detection}\label{sec:adv-algo}
A school of adversarial sample detection methods are based on confidence scores discrimination. By evaluating the confidence score based on training data density estimators in feature space, one can adequately judge whether a sample appears on the true data manifold with high probability or not.

\begin{figure}
    \centering
    \subfloat[\label{fig:baseline-features}]{%
        \includegraphics[width=0.5\columnwidth]{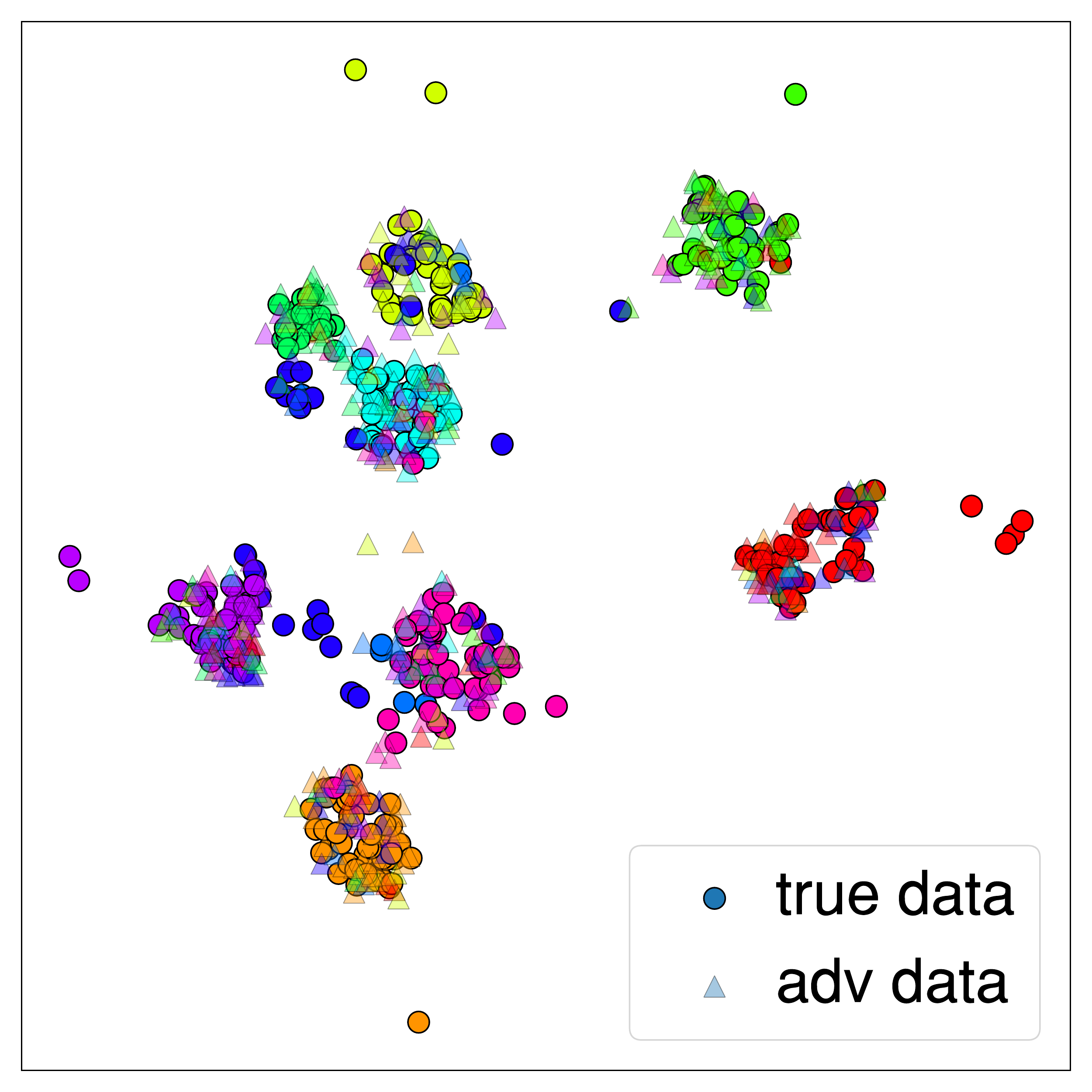}
    }
    \subfloat[\label{fig:lego-features}]{%
        \includegraphics[width=0.5\columnwidth]{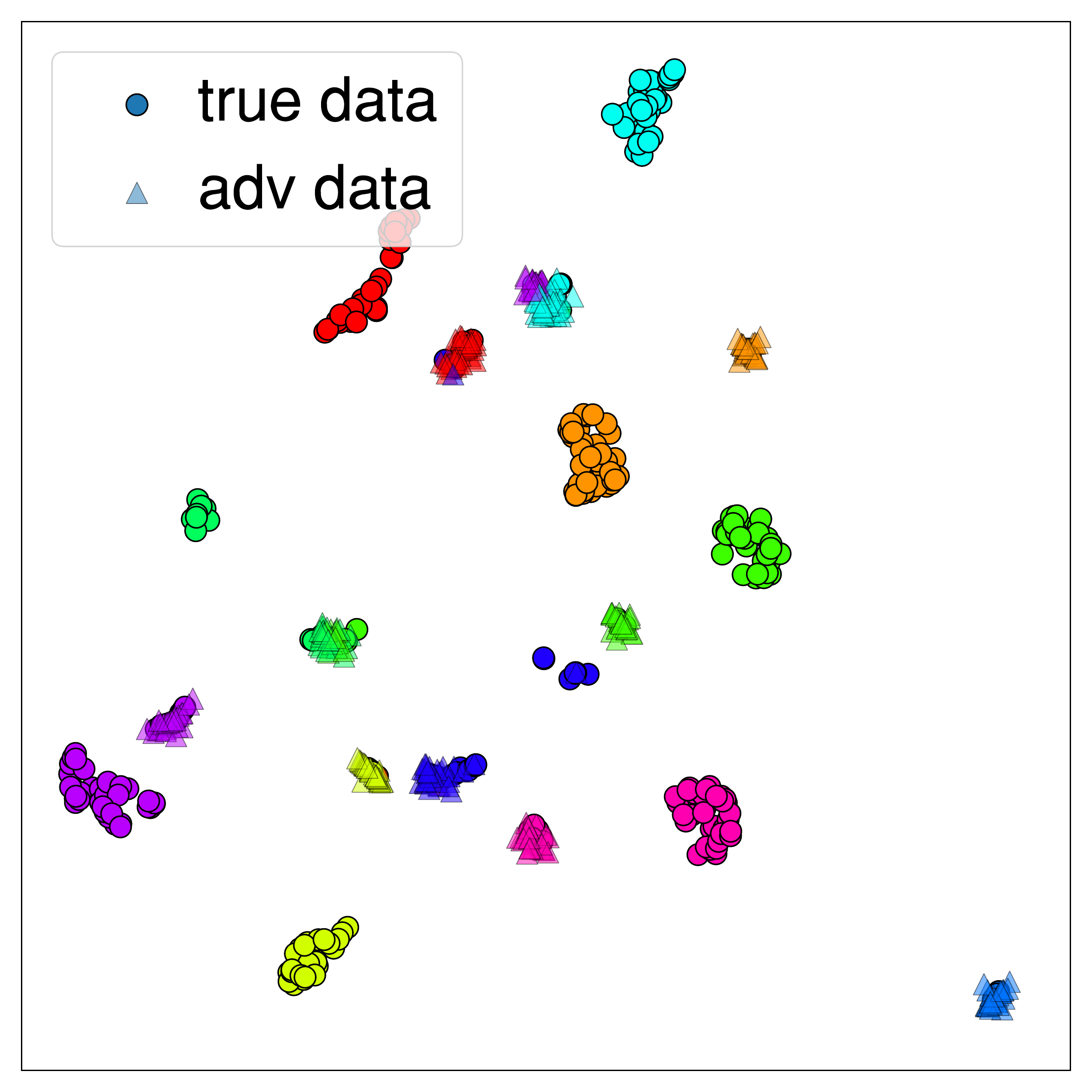}
    }
    \caption{Class-specific subnetworks can improve the discriminability between true data and adversarial data in feature space. Five hundred true images and corresponding adversarial samples are displayed by using the UMAP embedding projection of VGG16 penultimate features. The different predicted labels are marked with different colors. (a) feature embeddings generated by the original full model weights. (b) feature embeddings produced by their class-specific subnetworks.}
    \label{fig:adv-features}
    \vspace{-2ex}
\end{figure}

Now with the class-specific subnetworks, we can further increase the discriminability of confidence score methods.  Figure~\ref{fig:baseline-features} have displayed clustering patterns, but the adversarial samples have high overlaps with true data manifold. Figure~\ref{fig:lego-features} demonstrates that adversarial samples become more separable when using the features generated from class-specific subnetworks.

\begin{figure*}
    \centering
    \includegraphics[width=0.8\textwidth]{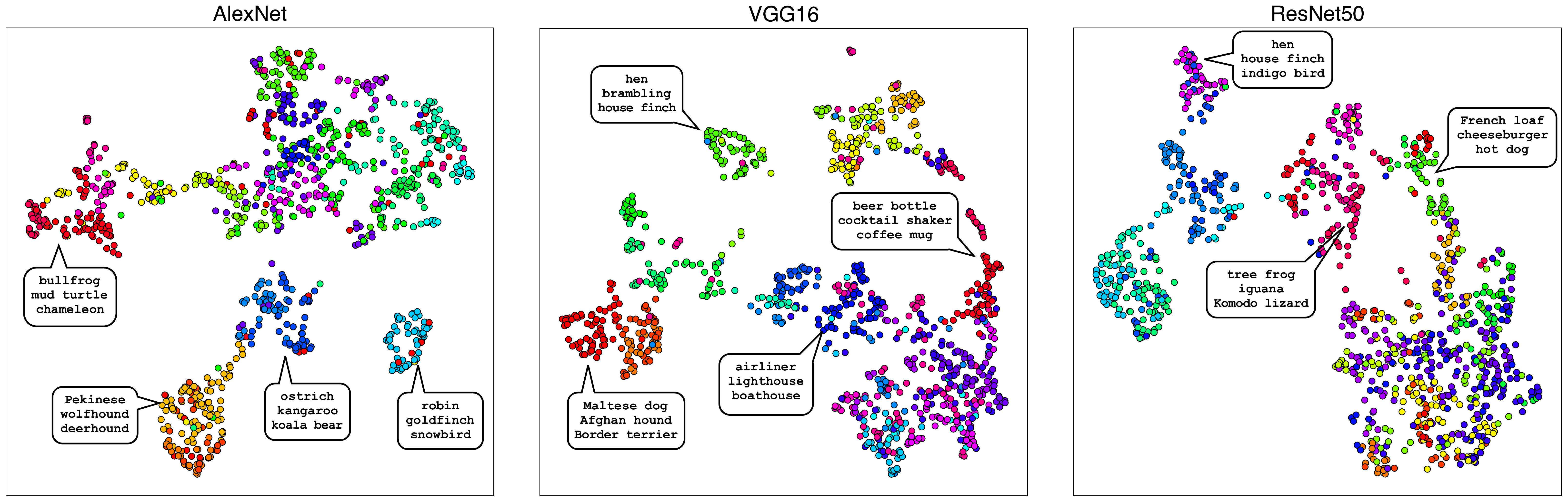}
    \caption{Subnetwork representation visualization by 2D UMAP embedding projection. For each figure, there show 1,000 subnetworks which are categorized into 25 clusters using a hierarchical clustering algorithm. For some apparent groups, we denote with their composition members' label names. These subnetwork representations display a resemblance to their class semantic similarity.  }
    \label{fig:umap} 
\end{figure*}

\begin{figure*}
    \centering
    \includegraphics[trim=0 400 0 0,clip,width=0.8\textwidth]{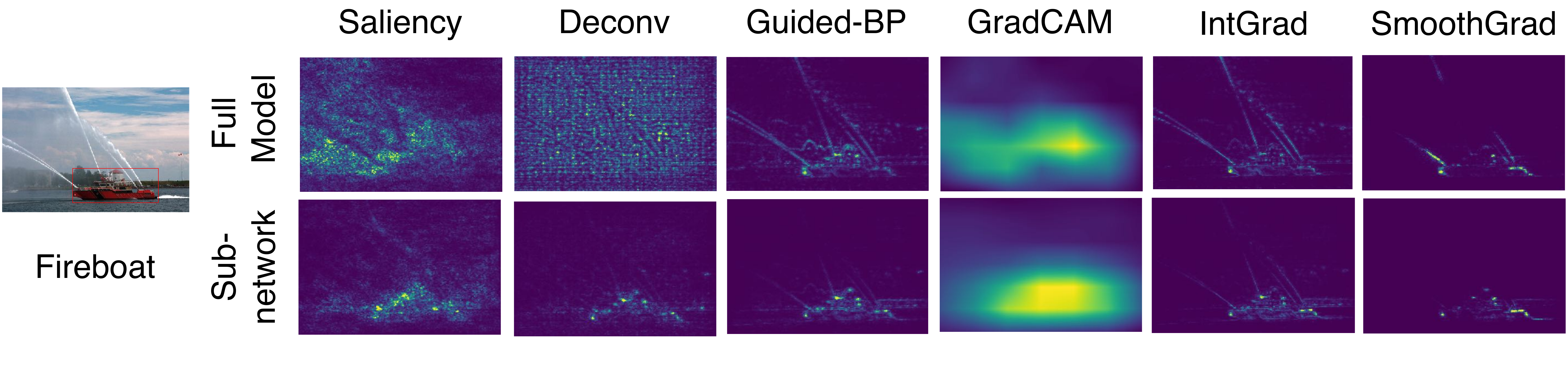}
    \caption{Visual explanation saliency by using class-specific model subnetworks. For each column, we show that for gradient-based explanation methods, the visual saliency can be improved by just replacing the full model weights with the subnetwork for the requested explanation class.}
    \label{fig:wsl}
\end{figure*}

\begin{table*}[!t]
  \centering
    \begin{tabular}{l|cc|cc|cc|cc|cc|cc}
    \toprule
    \multicolumn{1}{l}{\multirow{3}[3]{*}{Method}} & \multicolumn{4}{c}{AlexNet}   & \multicolumn{4}{c}{VGG16}     & \multicolumn{4}{c}{ResNet50} \\
    \multicolumn{1}{l}{} & \multicolumn{2}{c}{Normal} & \multicolumn{2}{c}{Subnet} & \multicolumn{2}{c}{Normal} & \multicolumn{2}{c}{Subnet} & \multicolumn{2}{c}{Subnet} & \multicolumn{2}{c}{Subnet} \\
\cmidrule{2-13}    \multicolumn{1}{l}{} & $\alpha^*$ & Err(\%) & $\alpha^*$ & Err(\%) & $\alpha^*$ & Err(\%) & $\alpha^*$ & Err(\%) & $\alpha^*$ & Err(\%) & $\alpha^*$ & Err(\%) \\
    \midrule
    Saliency & 3.5   & 46.11  & 3.5   & \textbf{44.80 } & 5.0   & 48.01  & 5.0   & \textbf{44.21 } & 5.0   & 49.20  & 4.0   & \textbf{41.64 } \\
    Deconv~\cite{simonyan2013deconv} & 4.5   & 49.61  & 4.5   & \textbf{47.00 } & 5.0   & 49.87  & 5.0   & \textbf{46.85 } & 3.5   & \textbf{47.26 } & 4.0   & 50.32  \\
    Guided-BP~\cite{springenberg2014guided} & 5.0   & 43.05  & 4.5   & \textbf{42.20 } & 7.0   & 41.66  & 6.0   & \textbf{40.83 } & 6.5   & 47.52  & 4.0   & \textbf{40.25 } \\
    GradCAM~\cite{selvaraju2017gradcam} & 1.0   & 54.13  & 1.0   & \textbf{51.48 } & 1.0   & 49.21  & 1.0   & \textbf{46.94 } & 1.0   & \textbf{42.63 } & 1.0   & 45.14  \\
     IntGrad~\cite{sundararajan2017intgrad} & 7.5	& 43.01	& 6.0 &	\textbf{42.71}	& 9.0	& 43.05	& 6.0	&\textbf{41.05}	& 8.5 & 47.60 & 4.0 & \textbf{41.85}  \\
     SmoothGrad~\cite{smilkov2017smoothgrad} & 7.0 & 48.72 & 4.0 & \textbf{47.01} &	9.5 & 53.57 & 2.0 & \textbf{47.75} & 3.0 & 52.37 & 2.5 & \textbf{51.98} \\
    \bottomrule
    \end{tabular}%
    \caption{Weakly supervised localization errors on ImageNet validation dataset. ``Normal'' indicates standard explanation practice by using full model weights regardless of the requested explanation class. ``Subnet'' indicates the proposed practice by replacing with class-specific subnetworks. $\alpha^*$ is optimized on held-out 5,000 images from ImageNet training dataset. ``Err'' indicates localization error (lower is better). }
  \label{tab:wsl}%
  
\end{table*}%

\begin{table*}
  \centering
  \caption{Adversarial sample detection AUROC (\%) for different methods. Our method improves upon Mahalanobis distance score by replacing feature extractor with class-specific subnetworks to estimate empirical means and covariance matrices. For unknown attack detection, FGSM samples denoted by ``seen'' are used for training logistic regression detector. }
    \begin{tabular}{cc|cccc|cccc}
    \toprule
    \multirow{2}[2]{*}{Dataset} & \multicolumn{1}{c|}{\multirow{2}[2]{*}{Mehod}} & \multicolumn{4}{c|}{Detection AUROC (\%)} & \multicolumn{4}{c}{Unknown Attack Detection AUROC (\%)} \\
          &       & FGSM  & BIM   & DeepFool & CW    & FGSM (seen) & BIM   & DeepFool & CW \\
    \midrule
    \multirow{4}[2]{*}{CIFAR-10} & KD + PU & 81.21 & 82.28 & 81.07 & 55.93 & 81.21 & 16.16 & 76.80 & 56.30 \\
          & LID   & 99.71 & 96.39 & 88.47 & 82.93 & 99.71 & 95.38 & 71.86 & 77.53 \\
          & Mahalanobis & 99.92 & \textbf{99.59} & 91.53 & 95.85 & 99.92 & 98.91 & 78.06 & 93.90 \\
          & Ours & \textbf{99.97} & 99.17 & \textbf{91.91} & \textbf{96.88} & \textbf{99.97} & \textbf{99.11} & \textbf{82.45} & \textbf{95.62} \\
    \midrule
    \multirow{4}[2]{*}{CIFAR-100} & KD+PU & 89.90 & 83.67 & 80.22 & 77.37 & 89.90 & 68.85 & 57.78 & 73.72 \\
          & LID   & 89.27 & 85.19 & 64.80 & 75.35 & 89.27 & 55.82 & 63.15 & 75.03 \\
          & Mahalanobis & 99.77 & 96.72 & \textbf{83.93} & 91.65 & 99.77 & \textbf{96.38} & \textbf{81.95} & 90.96 \\
          & Ours & \textbf{99.81} & \textbf{96.95} & 82.44 & \textbf{94.41} & \textbf{99.81} & 95.84 & 77.80 & \textbf{92.56} \\
    \midrule
    \multirow{4}[2]{*}{SVHN} & KD+PU & 82.67 & 66.19 & 89.71 & 76.57 & 82.67 & 43.21 & 74.26 & 67.85 \\
          & LID   & 95.72 & 87.41 & 88.81 & 85.66 & 95.72 & 84.88 & 67.28 & 76.58 \\
          & Mahalanobis & \textbf{99.63} & 97.14 & 95.46 & 92.13 & \textbf{99.63} & 95.39 & 72.20 & 86.73 \\
          & Ours & 99.54 & \textbf{97.24} & \textbf{95.82} & \textbf{93.63} & 99.54 & \textbf{96.38} &\textbf{78.75} & \textbf{91.09} \\
    \bottomrule
    \end{tabular}%
  \label{tab:adv}%
\end{table*}%

\noindent\textbf{Detection based on Mahalanobis distance} In this section, we will formally present the improved detection algorithm based on the Mahalanobis distance score proposed in~\cite{lee2018simple}. Following the definition in~\cite{lee2018simple}, the Mahalanobis confidence score $M(x)$ for a test sample $x$ is computed by measuring the Mahalanobis distance between $x$ and its
closest class-conditional Gaussian distribution in feature space:
\begin{equation}
    M(x) = \max_c - (f_\theta(x)-\hat{\boldsymbol{\mu}}_c)^\top\hat{\boldsymbol{\Sigma}}^{-1}(f_\theta(x)-\hat{\boldsymbol{\mu}}_c),
\end{equation}
where $\hat{\boldsymbol{\mu}}_c$ and $\hat{\boldsymbol{\Sigma}}$ are the empirical class mean and covariance for features of training samples $\{(x_1, y_1),\cdots,(x_N,y_N)\}$.

With the extracted subnetworks, we can modify the empirical mean and covariance estimation by using the class-specific feature as instead. Suppose that $\theta_c$ is the resulting subnetwork for class $c$. Then $\hat{\boldsymbol{\mu}}_c$ and $\hat{\boldsymbol{\Sigma}}$ can be estimated by
\begin{align}\label{eq:maha-score}
     \hat{\boldsymbol{\mu}}^{\mathrm{NEW}}_c = \frac{1}{N_c}&\sum_{i:y_i=c}f_{\textcolor{red}{\theta_c}} (x_i)\cr
    \hat{\boldsymbol{\Sigma}}^{\mathrm{NEW}} = \frac{1}{N}\sum_c\sum_{i:y_i=c}(f_{\textcolor{red}{\theta_c}} (&x_i)- \hat{\boldsymbol{\mu}}_c)(f_{\textcolor{red}{\theta_c}} (x_i) - \hat{\boldsymbol{\mu}}_c)^\top.
\end{align}

Similar to~\cite{ma2018characterizing}, the other low-level features in the neural network can also be combined to estimate confidence and a logistic regression detector is trained on a held-out validation data to weight each feature importance.

\section{Experiments}
We extract class-specific subnetworks of three typical ImageNet pre-trained networks: AlexNet, VGG16, and ResNet50. When optimizing the subnetwork for class $c$, for each epoch a balanced training set is sampled dynamically by including all the 1,000 images of class $c$, and an equal number of randomly chosen images for all the other classes. The final subnetwork is selected after $T=5$ epochs with minimum loss, below the sparsity level $\tau=0.5$. Balance parameter $\gamma=0.05$. Mini-batch size is 64. The learning rate for Adam optimizer is 0.1 for all the experiments.

\noindent\textbf{Subnetworks Visualization} For each subnetwork, the associated control gates can reflect the utilization of each layer when predicting specific class. Figure~\ref{fig:umap} displays the relationships between different class-specific subnetworks when projected onto the 2D plane using the UMAP algorithm. Here we can observe that the subnetwork representations tend to be more similar when their corresponding labels are semantically closer.

\noindent\textbf{Improving Visual Explanation} Visual explanation methods usually present the highlighted salient regions in the input image as explanation results. For most of visual explanation methods~\cite{simonyan2013deconv,springenberg2014guided,sundararajan2017intgrad,smilkov2017smoothgrad,selvaraju2017gradcam}, they all generate the visual saliency by following specific predefined  ``layerwise attribution backpropagation'' rules~\cite{MonDSP18}. Here we propose a simple alternative explanation procedure for the above gradient-based explanation methods, by \textit{using class-specific subnetwork as model weights} when explaining the requested class. Figure~\ref{fig:wsl} shows that by using the extracted class-specific subnetwork, these methods can generate more clear and accurate salient regions focusing on the main objects. 

\noindent\textbf{Weakly Supervised Object Localization} To demonstrate the improvement of visual explanation methods more rigorously, here we adopt the Weakly Supervised Object Localization (WSOL) evaluation protocol. Table~\ref{tab:wsl} summarizes the results. The proposed method can reduce localization errors across different methods. These results validate that the proposed practice can help improve gradient-based visual explanation methods.

\noindent\textbf{Detecting Adversarial Samples} Following the similar experimental setups in~\cite{lee2018simple}, we experiment with four attacking methods including FGSM, BIM, DeepFool and $\ell_2$-version CW attack. We first extract class-specific subnetworks for each dataset. Then the Mahalanobis scores are calculated according to Equation~\eqref{eq:maha-score}. The subsequent logistic regression detector setups are the same as~\cite{lee2018simple}.

We compared three state-of-the-art logistic regression detectors, which are based on 1) the combinations of kernel density (KD) and predictive uncertainty (PU)~\cite{feinman2017detecting}, 2) the local intrinsic dimensionality scores (LID)~\cite{ma2018characterizing} and 3) the Mahalanobis distance scores~\cite{lee2018simple}. 

The middle columns of Table~\ref{tab:adv} summarize the detection results. Our method can generally improve detection success rates over the baseline methods across different attacking methods. We also train the logistic regression detector on FGSM and evaluate its detection performance on the other types of adversarial samples. The right columns of Table~\ref{tab:adv} summarize the results. Our method can still outperform baseline methods in most cases. The results validate the power of class-specific subnetworks to detect adversarial examples.

\section{Conclusion}
In this paper, we explore the possibility of understanding DNNs from disentangled subnetworks. The discovery reveals that the extracted subnetworks can display a resemblance to their corresponding class semantic similarity. Furthermore, the proposed techniques can effectively improve the localization accuracy of visual explanation methods, and detection success rate of adversarial sample detection methods. 
{\small
\bibliographystyle{ieee}
\bibliography{egbib}
}

\end{document}